\documentclass[runningheads]{llncs}

\usepackage[final,year=2024,ID=2220]{eccv}

\usepackage{eccvabbrv}

\usepackage{graphicx}
\usepackage{booktabs}
\usepackage[pagebackref,breaklinks,colorlinks,citecolor=eccvblue]{hyperref}
\usepackage[accsupp]{axessibility}  

\usepackage{orcidlink}

\usepackage{comment}

\usepackage{graphicx}
\usepackage{subcaption}
\usepackage{float}

\usepackage{booktabs}                   
\usepackage{multirow}
\usepackage{array}
\usepackage{paralist}
\usepackage{enumitem}

\usepackage{mathtools}
\usepackage{amssymb,amsmath}   

\usepackage{pifont}
\usepackage{xspace}

\usepackage{diagbox}
\usepackage{xcolor}
\usepackage[export]{adjustbox}
\usepackage{sidecap}

\usepackage{multirow, makecell, booktabs}
\usepackage{wrapfig}
\definecolor{citecolor}{HTML}{0071bc}
\definecolor{linkcolor}{HTML}{ED1C24}

\def\eg{e.g.,~}               
\def\ie{i.e.,~}               

\newcommand{\figref}[1]{Figure~\ref{fig:#1}} 
\newcommand{\tabref}[1]{Table~\ref{tab:#1}}

\long\def\ignorethis#1{}

\newlength\paramargin
\newlength\figmargin
\newlength\subfigmargin
\newlength\secmargin
\newlength\subsecmargin
\newlength\tabmargin
\newlength\eqmargin

\newcolumntype{C}[1]{>{\centering\let\newline\\\arraybackslash\hspace{0pt}}m{#1}}

\newcommand*\colourmark[1]{%
  \expandafter\newcommand\csname #1xmark\endcsname{\textcolor{#1}{\ding{56}}}%
}

\colourmark{blue}
\colourmark{red}

\newcommand*\colourchecksnow[1]{%
  \expandafter\newcommand\csname #1snow\endcsname{\textcolor{#1}{\ding{100}}}%
}

\colourchecksnow{cyan}

\definecolor{darkspringgreen}{rgb}{0, 0.6, 0.3}
\newcommand*\colourcheck[1]{%
  \expandafter\newcommand\csname #1check\endcsname{\textcolor{#1}{\ding{52}}}%
}

\colourcheck{blue}
\colourcheck{cadmiumgreen}

\newcommand*\colourtri[1]{%
  \expandafter\newcommand\csname #1tri\endcsname{\textcolor{#1}{\ding{115}}}%
}

\colourtri{black}

\makeatletter
\@namedef{ver@everyshi.sty}{}
\makeatother
\usepackage{tikzsymbols}
\newcommand*\colourcheckfire[1]{%
  \expandafter\newcommand\csname #1fire\endcsname{\textcolor{#1}{\Fire}}%
}
\colourcheckfire{red}

\definecolor{cadmiumgreen}{rgb}{0.0, 0.42, 0.24}

\newcommand{\ourmodel}{AVSiam }
\newcommand{\ourmodela}{AVSiam}

\begin{document}

\title{Siamese Vision Transformers are \\ Scalable Audio-visual Learners}

\titlerunning{AVSiam}

\author{
Yan-Bo Lin \and
Gedas Bertasius
}
\authorrunning{Lin et al.}
%
%
\institute{Department of Computer Science \\
University of North Carolina at Chapel Hill\\
\email{\{yblin,gedas\}@cs.unc.edu}
}

\maketitle

\begin{abstract}
Traditional audio-visual methods rely on independent audio and visual backbones, which is costly and not scalable. In this work, we investigate using an audio-visual siamese network (\ourmodela) for efficient and scalable audio-visual pretraining. Our framework uses a single shared vision transformer backbone to process audio and visual inputs, improving its parameter efficiency, reducing the GPU memory footprint, and allowing us to scale our method to larger datasets and model sizes. We pretrain our model using a contrastive audio-visual matching objective with a multi-ratio random masking scheme, which enables our model to process larger audio-visual instance batches, helpful for contrastive learning. Unlike prior audio-visual methods, our method can robustly handle audio, visual, and audio-visual inputs with a single shared ViT backbone. Furthermore, despite using the shared backbone for both modalities, \ourmodela~achieves competitive or even better results than prior methods on AudioSet and VGGSound for audio-visual classification and retrieval. Our code is available at \url{https://github.com/GenjiB/AVSiam}
\end{abstract}  
\vspace{\secmargin}
\section{Introduction}\label{sec:intro}
\vspace{\secmargin}
The last few years have witnessed remarkable progress in audio-visual representation learning~\cite{av_iclr21_lee2021crossattentional,av_eccv20_avvp,eccv18_avel,av_cvpr21_av_parsing,cvpr22_ave_cmbs,cvpr22_avqa_avqa,cvpr22_ready_audio_adaptive}. However, most modern audio-visual approaches use separate audio and visual backbones and costly audiovisual pretraining protocols, limiting their scalability. For example, the recent state-of-the-art audio-visual method MAViL~\cite{nips23_mavil} requires $5,120$ V100 GPU hours for pretraining, which is not feasible for many research labs. Also, the best-performing variant of audio-visual MBT~\cite{nips21_bottleneck} uses more than 48GB of GPU memory during training, which requires costly A100 or H100 GPU servers, unavailable to many researchers. These factors make it difficult to scale these audio-visual models to larger datasets and bigger models, which has been the recent trend in many computer vision and multimodal modeling domains~\cite{arxiv23_onepiece,cvpr23_eva,tmlr22_coca,iclr23_pali}. 

In addition to large pretraining and GPU memory costs, the modality-specific backbones used by conventional audio-visual methods have several other shortcomings. First, designing separate audio and visual backbones may lead to hand-crafted priors and inductive biases, which not only contribute to increased research/engineering effort of optimizing models for a specific modality but can also be detrimental to data-driven representation learning~\cite{iccv23_bias_empirical,cvpr22_evading,eccv2022_equivariance}. Second, approaches relying on modality-specific backbones lack the flexibility to handle the cases of variable inputs and missing modalities (e.g., visual-only, audio-only, audio-visual, etc.). Lastly, the modality-specific audio-visual models are not parameter efficient, which increases their memory footprint and limits their scalability to larger datasets and bigger models.

Motivated by these observations, we introduce an Audio-visual Siamese network (\ourmodela) for efficient and scalable audio-visual pretraining. Our method is inspired by several recent works showing the ability of Vision Transformers (ViTs) to generalize to different modalities and domains, including audio and video~\cite{TASLP_AST,nips22_tvlt_textless,cvpr23_lavish,nips23_mavil,cavmae,avmae}. Also, since audio can be represented as a 2D spectrogram with a spatial 2D structure akin to \textit{audio images}, modern vision architectures (e.g., ViTs or CNNs) have been shown to process such audio images effectively. Despite these findings, most modern audio-visual methods~\cite{eccv22_eclipse,eccv22_joint_avvp,cvpr23_lavish,av_unified_icml23,nips23_mavil,cavmae,avmae} rely on modality-specific backbones for audio and visual data processing, since individually tailored models typically lead to better results on various audio-visual benchmarks.

\begin{figure}[t!]
    \centering
	\includegraphics[width=0.6\linewidth]{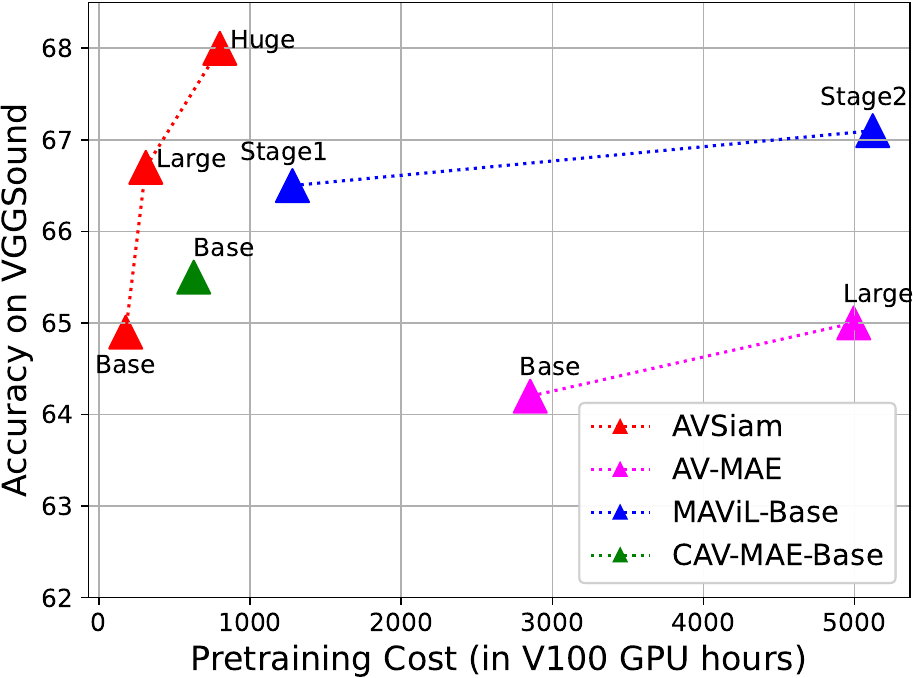}
    \caption{
    Our Audio-visual Siamese network (\ourmodela) uses a single shared backbone to process audio and visual data, which reduces its GPU memory footprint and allows us to scale our method to larger datasets and model sizes. Compared to prior audio-visual approaches~\cite{cavmae,nips23_mavil,avmae}, which are very costly, our model is both more efficient and also achieves higher accuracy on standard audio-visual classification benchmarks. \protect\footnotemark\vspace{-0.4cm}
    }
	\label{fig:teaser}
\vspace{\figmargin}
\end{figure}
\footnotetext{Since AV-MAE-Large did not report V100 hours, we approximate the pretraining hours based on our hardware.}

Instead, we investigate using a single Siamese vision transformer to process both audio and visual inputs, leading to a unified audio-visual model with a reduced GPU memory footprint. To pretrain our model, we use a contrastive audio-visual matching objective with a novel multi-ratio random masking scheme, which randomly masks audio and visual patches at different ratios.
Unlike fixed ratio masking~\cite{flip_cvpr23}, our multi-ratio random masking enables the model to learn robust representations across a spectrum of available information. 
%
Such a masking scheme and a shared audio-visual backbone also allow us to consider larger audiovisual instance batches, benefiting the contrastive learning process.

Our experimental results show that \ourmodela~can robustly process audio-only, visual-only, or audio-visual inputs with a single shared backbone. 
Furthermore, as depicted in~\figref{teaser}, despite using a shared backbone for both modalities, \ourmodel achieves competitive or superior results than prior methods with separate audio and visual backbones on audio-visual classification and retrieval benchmarks, including AudioSet-20K~\cite{icassp17_audioset}, AudioSet-2M~\cite{icassp17_audioset}, VGGSound~\cite{icassp20_vggsound}, and MSR-VTT~\cite{cvpr16_msrvtt} while requiring significantly fewer resources for pretraining (\ie $\bf28.9\times$ faster than MAViL)
%
Lastly, the efficiency of our approach enables training our model on larger datasets or with backbones of increased capacity, leading to further improvements. 

\vspace{\secmargin}
\section{Related Work}
\vspace{\secmargin}

\subsection{Audio-Visual Representation Learning}  
\vspace{\subsecmargin}
Audio-visual data provides a rich supervisory signal for learning audio and visual representations for the tasks of audio-visual event localization~\cite{eccv18_avel,wacv20_ave_avrb,icassp20_ave_avin,iccv19_ave_DAM,aaai20_ave_cman,eccv22_ave_DPNet,wacv23_AVE_CLIP}, audio-visual video parsing~\cite{av_cvpr21_av_parsing,av_eccv20_avvp,my_nips21,eccv22_joint_avvp,nips22_group_avvp}, and audio-visual classification~\cite{arxiv23_onepiece,nips23_mavil,cavmae,avmae,uavm,nips21_bottleneck,continue_av_iccv23}.
With the growing availability of audio-visual data on the Web, there has been a lot of progress in self-supervised audio-visual representation learning~\cite{av_iclr21_activeContrastive,av_cvpr21_RAVID,av_eccv18_Owens}.
The methods in~\cite{av_iccv17_look,av_eccv18_obj_that_sound,av_nips16_soundnet,av_eccv16_abSound,avt_nips20_VersatileNet,av_nips20_xdc,av_nips20_CrossLabelling,av_iclr21_activeContrastive,av_cvpr21_agreementAVID,av_cvpr21_RAVID} exploit the natural synchronization between audio and visual data by learning to predict whether a given audio and video pair matches.
Additionally, audio-visual contrastive learning methods~\cite{av_iclr21_activeContrastive,cav_temporal_aaai23,av_distilling_cl_cvpr21,nips21_av_contrastive} learn to associate audio-visual features from the same videos and differentiate representations from all other instances in a mini-batch.
Recently, masked autoencoders (MAE)~\cite{cavmae,nips23_mavil,avmae,nips22_tvlt_textless,av_speech_mae_iclr22,mae_ast_interspeech22,multimae_eccv22,expo_mae_arxiv22,nips22_audioMAE} have demonstrated the capability not only to learn robust visual features but also audio representations. 
Audio-visual MAE approaches~\cite{avmae,cavmae,nips23_mavil} reconstruct the original audio and visual data from masked audio-visual tokens to learn correlations between audio and images/video.
However, while effective, such MAE-based approaches~\cite{cavmae,nips23_mavil,avmae} typically require a high computational cost, which limits their scalability and adaptation.
To address these issues, we propose \ourmodela, an audio-visual model that uses a single backbone for audio and visual data to reduce the cost of audio-visual modeling while still attaining solid performance.

\begin{figure*}[t!]
    \centering
	\includegraphics[width=0.95\linewidth]{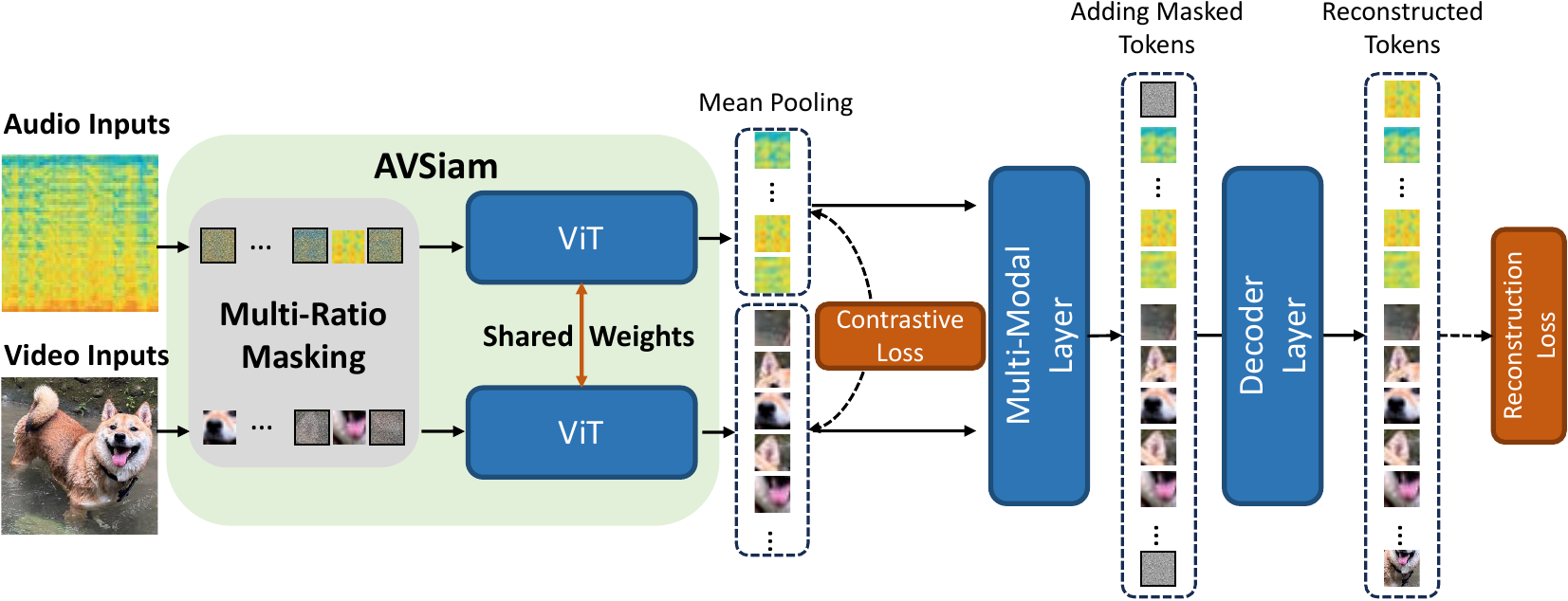}
\caption{\textbf{Our Pretraining Framework.} Our \ourmodela~approach uses a single shared vision transformer backbone to process both audio and visual data. To train our model, we use a novel multi-ratio masking scheme, which randomly masks audio and visual tokens at various masking ratios. As our pretraining objectives, we employ audio-visual contrastive matching and audio-visual token reconstruction loss functions. \vspace{-0.45cm}
}
    \vspace{\figmargin}
	\label{fig:method}
\end{figure*}

\subsection{Unified Multimodal Representation Learning} 
\vspace{\subsecmargin}
The versatile design of a Transformer~\cite{iclr21_vit} has made it easy to process multimodal data (\eg sound, images, videos, text, etc.), thus unlocking the potential for diverse multimodal applications~\cite{arxiv21_polyvit,zorro_arxiv23,my_accv20_av-trans,nips21_vatt,arxiv22_omnimae,cvpr22_omnivore,eccv22_eclipse}.
Several recent attention-based architectures~\cite{perceiver,perceiver_io,data2vec_icml22,clippo_cvpr23} have enabled direct processing of data across different modalities.
Another strategy for multimodal data processing involves augmenting models with modality-specific weights attached to a single modality-agnostic backbone~\cite{arxiv21_polyvit,nips21_vatt,one_model_arxiv22,arxiv_everyatonce}.
%
%
%
%
%
%
%
Furthermore, the models with shared weights across different tasks~\cite{moe_nips22,ms_clip,av_unified_icml23} recently achieved impressive results while demonstrating the flexibility of such approaches.
However, most existing audio-visual approaches~\cite{cavmae,nips23_mavil,avmae,nips21_bottleneck} still rely on modality-specific model design, which makes them costly and limits their scalability.
Instead, in this work, we introduce \ourmodela, a framework that uses a single audio-visual backbone for learning effective audio-visual representations in an efficient manner.

\vspace{\secmargin}
\section{Technical Approach}
\vspace{\secmargin}
\label{sec:method}

In this section, we present \ourmodela, an Audio-visual Siamese network that uses a shared-weight vision transformer (ViT) backbone to process audio and visual inputs for efficient audio-visual pretraining. We illustrate our pretraining framework in~\figref{method} and describe it in more detail below.
%
%
%

\subsection{The \ourmodela~Model}
\vspace{\subsecmargin}

\textbf{Audio-Visual Input Embeddings.} For the visual inputs, we use an RGB video frame $I \in \mathbb{R}^{H_v \times W_v \times 3}$, representing a frame randomly selected from the video at time $t$ with spatial dimensions $H_v \times W_v$. 
For the audio inputs, we process an audio spectrogram $A \in \mathbb{R}^{H_a \times W_a}$, covering approximately 10 seconds. %
Following ViT~\cite{iclr21_vit}, we patchify a RGB frame $I$ into $n$ non-overlapping patches, and transform them into visual embeddings $\mathbf{X}_v \in \mathbb{R}^{n \times d}$ 
Similarly, an audio spectrogram $A$ is projected into audio embeddings $\mathbf{X}_a \in \mathbb{R}^{k \times d}$ with $k$ patches. 
Note that due to different audio and visual channel dimensions, to obtain audio embeddings, we average the weights of the 3-channel projection layers of the pretrained ViT into single-channel weights when processing audio inputs.
\textbf{Model Architecture.} We use a standard ViT\cite{iclr21_vit} architecture to implement our \ourmodela~model. The main difference compared to prior transformer-based audio-visual approaches~\cite{cavmae, nips23_mavil, nips21_bottleneck} is that our model shares the full set of parameters of a pre-trained ViT across both audio and visual streams. 
We use a shared audio-visual encoder to process audio and visual inputs $\mathbf{X}_a$ and $\mathbf{X}_v$ and then pool the outputs for each modality via average pooling, which produces audio and visual features $\mathbf{F}_{a} \in \mathbb{R}^{d}$ and $\mathbf{F}_v \in \mathbb{R}^{d}$, respectively.
As for the multimodal layers and decoder, we follow the implementation of CAV-MAE~\cite{cavmae} and MAE~\cite{mae}. 
Specifically, the audio $\mathbf{F}_a$ and visual $\mathbf{F}_v$ tokens are concatenated and fed into a multimodal layer $\mathrm{MM}(.)$, which processes audio and visual tokens via a joint two-layer self-attention block.
During training, we also use a six-layer self-attention decoder $\mathrm{Dec}(.)$ to reconstruct the masked audio and visual tokens as shown in~\figref{method}.
We note that even though our \ourmodela~is based on the ViT design, in practice, it can be easily adapted to other backbones.

\begin{wrapfigure}{r}{0.49\textwidth}
    \vspace{-15mm}
    \centering
    \includegraphics[width=0.45\textwidth]{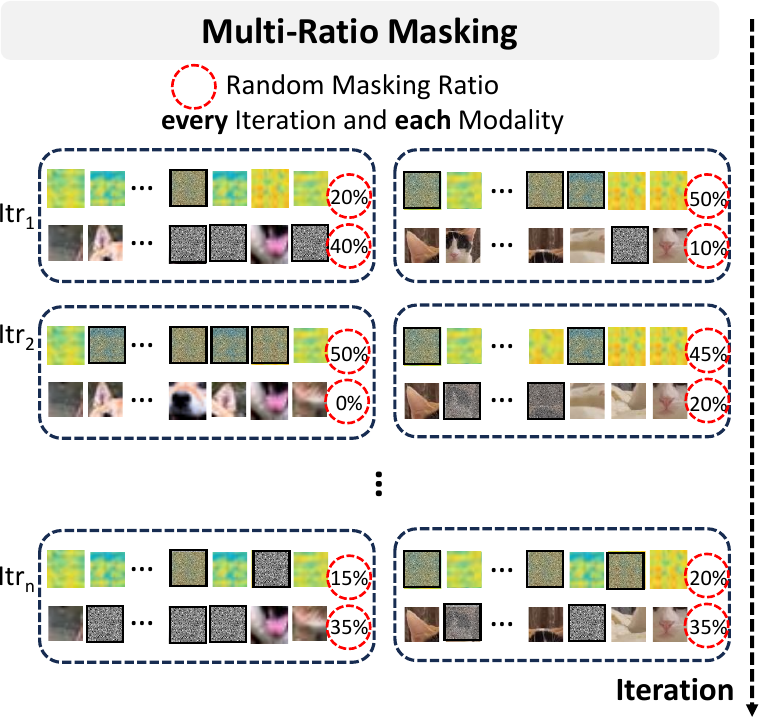}
    \caption{
        \textbf{Multi-Ratio Masking.} We apply random masking to audio and visual tokens in various proportions during each training iteration.
    }
    \label{fig:multi_ratio}
    \vspace{-8mm}
\end{wrapfigure}
\subsection{Training the \ourmodela~Model}
\vspace{\subsecmargin}

\textbf{Multi-Ratio Input Masking.} Prior work~\cite{flip_cvpr23,masked_features_cvpr22,mst_nips21} has shown that reducing the number of input tokens can lead to dramatic savings in computational resources (i.e., GPU memory usage, pretraining time, etc.). 
Since audio spectrogram inputs typically have substantially more tokens (patches) than images (\ie 512 vs. 196), such computational savings can be even more significant in the audio-visual domain. 
Although masking a large portion of tokens enables efficient large-scale pretraining, deciding a masking ratio that provides optimal efficiency vs. accuracy trade-off is not trivial.
For example, a high masking ratio can save GPU memory, leading to bigger batch sizes and, thus, potentially better performance for contrastive learning-based methods.
On the other hand, a high masking ratio also leads to a significant loss of information, which could negatively affect the quality of learned representations.

To incorporate the benefits of different masking ratios, we randomly mask audio and visual patches at varying ratios during each training iteration.
As depicted in \figref{multi_ratio}, during each pretraining iteration, every audio and visual instance in a mini-match will be randomly assigned a masking ratio from 0\% to 50\%. 
Unlike the fixed ratio masking approach~\cite{flip_cvpr23}, which always uses the same number of tokens, our multi-ratio random masking scheme learns from a diverse number of tokens, which leads to more robust audio-visual representations. 
To efficiently implement our proposed masking scheme, we select the ratios from a predefined set of discrete ratios for each iteration (randomly), rather than by randomly selecting each ratio in a batch. This ensures that the number of instances with an identical number of unmasked tokens is predictable and can be stacked together for efficient GPU memory utilization.%

\textbf{Audio-visual Pretraining Objectives.} 
We use audio-visual contrastive matching and token reconstruction
objectives to pretrain our model. Specifically, we use audio $\mathbf{F}_a$ and visual features $\mathbf{F}_v$, obtained from our shared \ourmodela~encoder for contrastive audio-visual matching as in~\cite{infonce,cavmae}:
\begin{equation}
\begin{aligned}
\label{eq:contrastive}
\mathcal{L}_{c}(\mathbf{F}_a,\mathbf{F}_v) = -\frac{1}{B} \sum_{i=1}^B {\rm log}  \frac{ {\mathrm exp} (\mathrm{g}(\mathbf{F}_{a}^{i},\mathbf{F}_{v}^{i})/\tau)} {\sum_{j=1}^{B} {\mathrm exp} (\mathrm{g}(\mathbf{F}_{a}^{i},\mathbf{F}_v^{j})/\tau)) }.
 \end{aligned}
\end{equation}
Here, $\mathrm{g(.)}$ is a standard cosine similarity function, and $B$ and $\tau$ denote mini-batch size and temperature, respectively.
Also, inspired by the recent audio-visual MAE approaches~\cite{cavmae, avmae, nips23_mavil}, we adopt a masked token reconstruction~\cite{mae} objective to learn a more effective audio-visual representation.  Specifically, the audio-visual decoder $\mathrm{Dec(.)}$ processes the masked and unmasked audio and visual tokens as inputs to predict the original audio spectrogram and image as: 

\begin{equation}
\begin{aligned}
\label{eq:rec}
\tilde{I}, \tilde{A} = {\mathrm Dec}(\mathrm{MM}(\mathbf{F}^{u}_a,\mathbf{F}^{u}_v),\mathbf{F}^{m}_a,\mathbf{F}^{m}_v),
\end{aligned}
\end{equation}

where $\mathbf{F}^{u}_a, \mathbf{F}^{u}_v, \mathbf{F}^{m}_a$, and $\mathbf{F}^{m}_v$ are unmasked audio, unmasked visual, masked audio, and masked visual tokens respectively.
The reconstructed audio spectrogram $\tilde{A} \in \mathbb{R}^{H_a\times W_a} $ and image $\tilde{I} \in \mathbb{R}^{H_v \times W_v \times 3}$ has the same shape as the original spectrogram and image inputs.
The reconstruction objective is written as:
\begin{equation}
\begin{aligned}
\label{eq:rec_loss}
\mathcal{L}_{rec} &= \frac{1}{B} \sum_{i=1}^{B} (\tilde{A}^{i}- A^{i})^2 + (\tilde{I}^{i}- I^{i})^2.
 \end{aligned}
\end{equation}
The final objective is obtained by combining the two losses: $\mathcal{L} = \mathcal{L}_{rec}+\mathcal{L}_{c}$. %

\textbf{Supervised Finetuning with Mixed Modality Inputs.}  After pretraining our framework as described above, we proceed with supervised fine-tuning of our model on various downstream audio-visual understanding tasks.
Unlike in the pretraining stage, in the supervised fine-tuning stage, we do not mask any of the input tokens and use a standard training protocol as in CAV-MAE~\cite{cavmae} and MAViL~\cite{nips23_mavil}.
To robustly handle missing modalities and ensure generalization to audio-only and visual-only settings, during each iteration of finetuning, we randomly select one of three types of inputs, audio-only, visual-only, or audio-visual, and feed it into our shared ViT encoder.
Once our shared encoder processes these inputs, we average all tokens encoded by the ViT to obtain representations for each modality (i.e., $\mathbf{F}_a$ for audio-only, $\mathbf{F}_v$ for visual-only, $\mathbf{F}_a$ and $\mathbf{F}_v$ for audio-visual inputs).
The resulting features are then fed into an MLP layer for the final prediction. 
We use binary cross-entropy loss for AudioSet~\cite{icassp17_audioset} and cross-entropy loss for VGGSound~\cite{icassp20_vggsound} as in CAV-MAE~\cite{cavmae} for our supervised finetuning. 

\subsection{Implementation Details}
\vspace{\subsecmargin}
\label{sec:imp}
We used a standard Vision Transformer (ViT) architecture for the base, large, and huge variants.
All encoder layers, adapted from ViT, were pretrained on ImageNet-21K. %
%
%
%
%
%
For the multimodal layers, 
we utilized the weights from the last two layers of the pretrained ViT to initialize the first two layers of the decoder to process all unmasked audio-visual tokens.
As for the decoder, we adhered to the default MAE settings for the base, large. Specifically, the decoder processed all masked and unmasked tokens as inputs where the masked tokens were randomly initialized and learnable. The weights of the decoder were randomly intitialized.
For contrastive loss, the temperature was set to \(\tau=0.05\).
We utilized two independent loss scalers, one for contrastive loss and the other for the MAE loss.
During supervised finetuning with mixed modality inputs (see details above), we randomly sampled either audio or visual inputs in half of our training iterations to train our model on audio-only or visual-only inputs. In the remaining iterations, we used both audio and visual inputs. 
As for optimization, we used Adam optimizer for both the pretraining and fine-tuning stages. 
We set the learning rate to $1e-4$ for pretraining. During finetuning, we set $1e-4$, $5e-6$, and $5e-5$ for AudioSet-20K, AudioSet-2M, and VGGSound. 
%
%
%

\vspace{\secmargin}
\section{Experimental Setup}
\vspace{\secmargin}
\label{sec:exp_setup}

\indent\textbf{Audio-visual Classification.} We pretrain our model on AudioSet-2M~\cite{icassp17_audioset}. We note that we could only obtain 1.7M videos due to expired YouTube links.
After pretraining, the model is finetuned on three audio-visual classification datasets: (1) AudioSet-20K containing 20K samples from the same domain as the pretraining data (i.e., AudioSet-2M), (2) the original AudioSet-2M dataset, and (3) VGGSound~\cite{icassp20_vggsound}. 
%
For evaluation metrics, we use standard mean Average Precision (mAP) for AudioSet-20K and AudioSet-2M and top-1 accuracy for VGGSound.
%
%

\textbf{Audio-visual Retrieval.} 
To evaluate whether our learned audio and visual representations generalize to different tasks, we also evaluate \ourmodel on video-to-audio and audio-to-video retrieval tasks without any additional finetuning similar to the setting of CAV-MAE~\cite{cavmae}.
Additionally, we follow the setup of MAViL~\cite{nips23_mavil}, and also include audio-to-video retrieval results after finetuning the pretrained model. %
The recall at 1  metric (\(R@1\)) is then computed using the cosine similarity between the audio and visual representations. 
We use the original evaluation splits from CAV-MAE, which include 1,725 and 1,545 videos from AudioSet and VGGSound, respectively. We also include audio-visual retrieval results on the test set of MSR-VTT~\cite{cvpr16_msrvtt} containing 2,990 videos.

\vspace{\secmargin}
\section{Results and Analysis}
\vspace{\secmargin}
\label{sec:results}
We first discuss our audio-visual classification results on AudioSet and VGGSound. Afterward, we present our analysis for audio-visual retrieval on AudioSet, VGGSound, and MSR-VTT. Lastly, we discuss our ablation studies.

\begin{table*}[t]
\caption{
    \textbf{Audio-Visual Event Classification:} 
    We compare the results of our proposed \ourmodel with previous baselines on audio-visual event classification on the AudioSet and VGGSound benchmarks.
    We de-emphasize MBT*~\cite{nips21_bottleneck} due to using non-standard training-test splits.
    Note that \ourmodela-Base$^{+}$ uses more data from ACAV~\cite{acav100m} dataset for pretraining only.
    }
\vspace{\tabmargin}
\centering
\resizebox{0.95\linewidth}{!}{
\begin{tabular}{lccccccc}
Method    & \makecell{Audio Encoder} & \makecell{Visual Encoder} & \makecell{V100 \\ Hours}  &\makecell{\#Param }   & \makecell{AS-20K \\(mAP$\uparrow$)} & \makecell{AS-2M \\(mAP$\uparrow$)} & \makecell{VGGSound \\ (Acc.$\uparrow$)} \\ \midrule

\multicolumn{8}{l}{\textit{\textbf{Audio-Video Models}}}\\
G-Blend~\cite{gblend} & - & -& - &  -  & 37.8 & 41.8 & - \\
Perceiver~\cite{perceiver}  & -& -& -&  -  & - & 44.2 & - \\
Attn AV~\cite{attnav_ijcai20}  & -& - & - & -  & -  & 44.2 & - \\
Zorro-Swin~\cite{zorro_arxiv23}  & -& - &  - & 161M & - & 46.5& - \\
\color{gray}MBT\textsuperscript{*}{\cite{nips21_bottleneck}} & \color{gray}AST-B & \color{gray}ViT-B& -   & \color{gray}172M  & \color{gray}43.9  & \color{gray}49.6 & 64.1 \\
\midrule
\multicolumn{8}{l}{\textit{\textbf{Masked Autoencoder}}}   \\
AV-MAE~\cite{avmae}               & AST-B& ViT-B& 2854 &179M    & -  & 50.0 & 64.2 \\
AV-MAE-Large   & AST-L& ViT-L & -&626M   & -  & 51.8 & 65.0 \\
CAV-MAE~\cite{cavmae}      & AST-B& ViT-B& 672 &164M    & 42.0  & 51.2  & 65.5 \\
MAViL-Stage1~\cite{nips23_mavil} & AST-B& ViT-B& 1280 &172M   & 44.6  & 51.9 & 66.5  \\      
MAViL-Stage2 & AST-B& ViT-B& 5120&172M    & 44.9 & 53.3  & 67.1  \\   

\midrule
\multicolumn{8}{l}{\textit{\textbf{Shared-Weight Encoder}}}  \\
\ourmodela-Base  & \multicolumn{2}{c}{ViT-B (shared)}& 177  & \multicolumn{1}{c}{\bf100M} & 41.6  & 50.1 & 64.9 \\ 
\ourmodela-Base$^{+}$  & \multicolumn{2}{c}{ViT-B (shared)}& 450  & \multicolumn{1}{c}{\bf100M} & 43.0  & 51.4    & 66.7 \\ 
\ourmodela-Large  & \multicolumn{2}{c}{ViT-L (shared)}& 310 & \multicolumn{1}{c}{332M}  & 44.1 & 52.1  & 67.1 \\ 
\ourmodela-Huge  &  \multicolumn{2}{c}{ViT-H (shared)}& 800 & \multicolumn{1}{c}{672M} & \textbf{45.0}& \textbf{54.1}  & \textbf{68.0} \\ 
\label{tab:sota}
\end{tabular}
}
\vspace{-0.5cm}
\end{table*}

\subsection{Audio-Visual Classification Results}
\vspace{\subsecmargin}

In \tabref{sota}, we present our audio-visual classification results on AudioSet and VGGSound using multiple axes of comparison, including the types of audio and visual encoders, the pretraining time in V100 GPU hours, the number of parameters, and audio-visual classification accuracy. 
We compare our \ourmodel~approach (bottom part of the table) with standard audio-visual models~\cite{gblend,perceiver,attnav_ijcai20,zorro_arxiv23,nips21_bottleneck} (top part of the table) and also with recent MAE-based approaches~\cite{avmae,cavmae,nips23_mavil} (middle part of the table).
Based on these results, first, we observe that our default \ourmodela-Base model performs similarly as AV-MAE~\cite{avmae} and MBT~\cite{nips21_bottleneck} on AudioSet-2M (\ie \textbf{50.1} mAP vs. \textbf{50.0} mAP vs. \textbf{49.6} mAP) and VGGSound (\ie  \textbf{64.9\%} vs. \textbf{64.2\%} vs. \textbf{64.1\%}). 
We also observe that \ourmodel~achieves worse results than the latest state-of-the-art MAE-based approaches such as CAV-MAE~\cite{cavmae} and MaViL~\cite{nips23_mavil} on AudioSet-2M (\ie \textbf{50.1} mAP vs. \textbf{51.2} mAP vs. \textbf{53.3} mAP) and VGGSound (\ie \textbf{64.9\%} vs. \textbf{65.5\%} vs. \textbf{67.1\%}.
However, unlike all these other approaches, which use modality-specific audio and visual encoders, \ourmodela~uses a single, shared audio-visual backbone, which has a smaller number of parameters  (\ie \textbf{100M} vs. \textbf{164M} vs. \textbf{172M}) and requires significantly less time for pretraining (\ie \textbf{177} vs. \textbf{672} vs. \textbf{5120} V100 GPU hours.
In particular, we note that the best performing MaViL variant requires \textbf{5,120} V100 GPU hours for pretraining which is $\bf28.9\times$ more than our model (\ie \textbf{177} V100 GPU hours).
These results highlight the efficiency of our approach.

Next, leveraging the efficiency of our model, we show that we can scale \ourmodela~to larger training datasets and also larger model variants (i.e., ViT-Large, ViT-Huge).
Specifically, we pretrain our model on AudioSet-2M~\cite{icassp17_audioset} augmented with additional samples from VGGSound (200K)~\cite{icassp20_vggsound} and ACAV (2.4M)~\cite{acav100m}. %
We report these results under the \ourmodela-Base$^{+}$ variant. 
We observe that scaling our model to a larger dataset size consistently improves the performance across all three datasets (\ie \textbf{+1.4\%}, \textbf{+1.3\%}, and \textbf{+1.8\%}).
Moreover, \ourmodela-Base$^{+}$ outperforms the most efficient MAE baseline, CAV-MAE (\ie AudioSet-20K: \textbf{43.0\%} vs. \textbf{42.0\%}, AudioSet-2M: \textbf{51.4\%} vs. \textbf{51.2\%}, and VGGSound: \textbf{66.7\%} vs. \textbf{65.5\%}) while still being faster to pretrain (\ie \textbf{450} vs. \textbf{672} V100 hours) despite using a significantly larger dataset size.

Afterward, we also investigate scaling our model size by considering the \ourmodela-Large and \ourmodela-Huge variants. %
Compared to \ourmodela-Base, \ourmodela-Large and \ourmodela-Huge demonstrate consistent and significant improvements in audio-visual classification on AudioSet-20K (\textbf{+2.5} mAP and \textbf{+3.4} mAP), AudioSet-2M (\textbf{+2} mAP and \textbf{+4.1} mAP), and VGGSound (\textbf{+2.2\%} and \textbf{+3.1\%}), respectively. 
It is worth pointing out that training the largest variant of our model, \ourmodela-Huge, is still a lot cheaper than training any of the AV-MAE, MAViL-Stage1, MAViL-Stage2 baselines (\ie \textbf{800} vs. \textbf{2854}, \textbf{1280}, and \textbf{5120} V100 hours).
Moreover, the improvements with our model from Base to Large are greater than those seen in AV-MAE~\cite{avmae}, with gains on AudioSet-2M (\textbf{+2 mAP}  vs. \textbf{+1.8 mAP} ) and VGGSound (\textbf{+2.2\%} vs. \textbf{+0.8\%}). 
These results demonstrate that \ourmodel scales well both with respect to the training dataset size as well as the model size. 
We also note that our largest variant, \ourmodela-Huge, achieves the state-of-the-art results on AudioSet-20K~\cite{icassp17_audioset} (\textbf{45.0} mAP), AudioSet-2M~\cite{icassp17_audioset} (\textbf{54.1} mAP), and VGGSound~\cite{icassp20_vggsound} (\textbf{68.0\%} accuracy) while also requiring only \textbf{15\%} of pretraining time of the previous best-performing method, MAViL-Stage2.
\begin{table*}[t]
\centering
\caption{\textbf{Zero-shot Audio-Visual Retrieval:}
We evaluate \ourmodel for video-to-audio and audio-to-video retrieval on the AudioSet, VGGSound, and MSR-VTT datasets, all in zero-shot settings.
The results are reported using the recall at 1 metric (R@1).
Compared to prior approaches, \ourmodel achieves the best results across all metrics and datasets while using fewer parameters. 
}
\vspace{\tabmargin}
\resizebox{0.95\linewidth}{!}{
\begin{tabular}{@{}lccccccccc@{}}
\toprule
& \multirow{2}{*}{\makecell{Audio \\ Encoder}}  &\multirow{2}{*}{\makecell{Visual \\ Encoder}} &\multirow{2}{*}{\makecell{\#Params (M)}}  & \multicolumn{2}{c}{AudioSet}      & \multicolumn{2}{c}{VGGSound}    & \multicolumn{2}{c}{MSR-VTT}   \\ 
                                    &&&& V$\rightarrow$A           & A$\rightarrow$V           & V$\rightarrow$A          &  A$\rightarrow$V            &  V$\rightarrow$A           &  A$\rightarrow$V           \\ 
\midrule
%
CAV-MAE                                 &AST-B&ViT-B& 164& 16.1&13.5                          & 14.7&12.1        & 4.9       & 8.3         \\
CAV-MAE\textsuperscript{+}              &AST-B&ViT-B& 164& 18.8&15.1                           & 14.8&12.8          & 7.6          & 13.3          \\ \midrule
\ourmodela-Base                                         &\multicolumn{2}{c}{ViT-B (shared)}& \bf100& \bf19.7&\bf17.6   & \bf19.0&\bf20.4          &   \bf9.3      &   \bf16.1        \\ \bottomrule
\label{tab:sota_retrieval}
\end{tabular}
}
\vspace{-0.4cm}
\end{table*}

\subsection{Audio-visual Retrieval Results}
\vspace{\subsecmargin}
In \tabref{sota_retrieval}, we compare \ourmodel with two variants of the CAV-MAE~\cite{cavmae} on video-to-audio and audio-to-video retrieval on AudioSet, VGGSound and MSR-VTT.
We note that the CAV-MAE$^{+}$ baseline is trained with a larger batch size than the standard CAV-MAE variant. 
Our results indicate that for video-to-audio retrieval, \ourmodela-Base outperforms both CAV-MAE variants on AudioSet (\ie \textbf{19.7} vs. \textbf{18.8}), VGGSound  (\ie \textbf{19.0} vs. \textbf{14.8}) and MSR-VTT   (\ie \textbf{9.3} vs. \textbf{7.6}) while requiring significantly fewer parameters (\ie \textbf{100M} vs. \textbf{164M}). 
We also observe that the performance gap of our model w.r.t CAV-MAE variants is even larger for audio-to-video retrieval where \ourmodela-Base outperforms CAV-MAE$^{+}$ in AudioSet (R@1: \textbf{17.6} vs. \textbf{15.1}), VGGSound (R@1: \textbf{20.4} vs. \textbf{12.8}) and MSR-VTT (R@1: \textbf{16.1} vs. \textbf{13.3})
Based on these results, we hypothesize that using a single shared audio-visual backbone is beneficial for audio-vsual retrieval tasks compared to using separate audio and visual encoders. This is because a shared encoder projects audio and visual inputs into a common latent representation space, which may be beneficial for retrieving the correct videos based on the audio inputs and vice-versa.
%
%
%
%
We also note that the performance gap between \ourmodel and CAV-MAE\textsuperscript{+} is bigger on VGGSound than on AudioSet and MSR-VTT. 
This is because VGGSound contains higher-quality audio-visual pairs, which makes it more suitable for evaluating video-to-audio retrieval. Overall, despite using significantly fewer parameters, \ourmodel outperforms all competing approaches by a large margin on all audio-visual retrieval benchmarks.

\textbf{Additional Results on MSR-VTT.} In addition to our zero-shot audio-visual retrieval results, in \tabref{rebuttal_msr_vtt_ft}(a), we also include the results after finetuning on MSR-VTT as was done in~\cite{nips23_mavil}. We then compare our \ourmodela-Base variant with the state-of-the-art MAViL baseline. Note that we follow the evaluation protocol of MaViL~\cite{nips23_mavil}, and only report audio-to-video retrieval results using the R@1 metric. Based on these results, we observe that \ourmodela-Base achieves better audio-to-video retrieval results than MAViL (\textbf{24.3} vs. \textbf{22.8}).

\begin{table}[t]
    \centering
    \caption{Evaluations of \ourmodela~against MAViL~\cite{nips23_mavil} after both models were finetuned on MSR-VTT~\cite{cvpr16_msrvtt} for audio-to-video retrieval (\textbf{left}) and comparisons in throughput performance (\textbf{right}). All models are implemented using the same GPU hardware.}
    \begin{minipage}{0.46\linewidth}
        \centering
        \caption*{(a) A2V Retrieval on MSR-VTT.
        %
        %
        }
            \label{tab:rebuttal_msr_vtt_ft}
            \centering
            \resizebox{0.99\linewidth}{!}{
            \begin{tabular}{l cc}
                \toprule
                {Method}  & MAViL & \ourmodela-Base \\  
                \midrule
                R@1 Accuracy        &22.8  &\bf24.3  \\
                \bottomrule
            \end{tabular}
            }
    \end{minipage}%
    \hfill
    \begin{minipage}{0.54\linewidth}
       \centering
       %
        \caption*{
        (b) Throughput Comparison.
        }
        \label{tab:rebuttal_flops}
        \centering
        \resizebox{0.99\textwidth}{!}{
        \begin{tabular}{l ccc}
            \toprule
            {Method} &  MAViL & CAV-MAE &\ourmodela \\
            \midrule
            Samples/Sec. $\uparrow$ & 3.84 & 22.5  &    \bf75.4 \\
            \bottomrule
            
        \end{tabular}
        }
    \end{minipage}
    \vspace{-0.3cm}
\end{table}

\begin{table}[t]
    \vspace{+2mm}
    \caption{
    \textbf{Multi-ratio Masking vs. Fixed-ratio Masking.}
    We compare our multi-ratio input masking scheme with various fixed-ratio masking schemes as well as the contrastive learning baseline. We use audio-visual classification accuracy and efficiency in terms of V100 GPU pretraining hours as our metrics of comparison.
    All variants are pretrained on AudioSet-2M and finetuned on AudioSet-20K. 
    }
    \vspace{\tabmargin}
    \label{tab:ratio}
    \centering
    \resizebox{0.65\linewidth}{!}{
    \begin{tabular}{l|ccc|cccc}
        \multicolumn{1}{l}{{{}}}& \multicolumn{6}{c}{AS-20K (mAP)$\uparrow$}  \\
        \toprule
          \diagbox[width=12em]{A Ratio}{V Ratio}&  \multicolumn{3}{c|}{Acc.$\uparrow$}    &  \multicolumn{3}{c}{V100 Hours $\downarrow$}    \\
                                       &25\%  &50\%  &75\%    & 25\%& 50\% & 75\% \\
        \midrule
            \makecell{25\%}                              &40.8   &40.3  &40.0    & 362& 326 & 300 \\
            \makecell{50\%}                              &39.8   &39.5  &  39.4  & 163& 142& 136 \\
            \makecell{75\%}                              &39.0   &38.8  & 38.6   & 138& 132& \bf120 \\
            \midrule
           Contrastive Learning                               & \multicolumn{3}{c|}{40.4} & \multicolumn{3}{c}{510} \\
            Multi-ratio Masking (Ours)                                & \multicolumn{3}{c|}{\bf41.3} & \multicolumn{3}{c}{160}\\
        \bottomrule
    \end{tabular}
}
\vspace{\tabmargin}
\end{table}

\subsection{Throughput Comparison}
\vspace{\subsecmargin}

In \tabref{rebuttal_flops}(b), we also include throughput comparisons of our method and the two best performing audio-visual approaches, CAV-MAE and MaViL. The throughput is measured using the number of processed samples per second. All methods are implemented using identical GPU hardware (\ie single NVIDIA A5000).
Our results indicate that \ourmodela-Base achieves the highest samples per second throughput (\textbf{75.4}), outperforming both CAV-MAE (\textbf{22.5}) and MAViL-Stage2 (\textbf{3.84}). This suggests that our shared audio-visual encoder design is helpful for improving not only the parameter efficiency but also its throughput.
\begin{table}[t]
\caption{
     \textbf{Comparison with a Separate-encoder Baseline.} We compare our \ourmodel~with the AVSep baseline that uses separate audio and visual encoders on AudioSet-2M. AVSiam achieves similar or even better performance than AVSep while being significantly more parameter-efficient and requiring \textbf{1.9$\times$} less GPU memory.
}
\vspace{\tabmargin}
    \label{tab:abs_gpu}
    \centering
    \setlength{\tabcolsep}{6pt} 
    \resizebox{0.7\textwidth}{!}{
    \begin{tabular}{l cccccc}
      \toprule
        Method            &\makecell{\#Param} $\downarrow$& A &V&A+V & GPU Mem.\\
        
        \midrule
          AVSep-Large                                        & 640M   & \bf48.0 &29.8& 52.0 & 20.6G     \\
          AVSiam-Large                                       & \bf332M   & 47.8 & \bf30.1 & \bf52.1  & \textbf{10.9G}     \\
        \bottomrule
    \end{tabular}
}
\vspace{\tabmargin}
\vspace{-0.3cm}
\end{table}

\begin{table*}[t]
\caption{
    \textbf{Comparing ViT vs. AST encoders.} We compare our default method that uses ViT as its shared audio-visual encoder with a variant that uses an Audio Spectrogram Transforme AST~\cite{interspeech21_AST} instead of a ViT. We report that using ViT as our shared audio-visual encoder produces significantly better results in visual-only and audio-visual classification settings.
}
\vspace{\tabmargin}
    \label{tab:rebuttal_ast}
    \centering
    \setlength{\tabcolsep}{6pt} 
    \resizebox{0.65\linewidth}{!}{
    \begin{tabular}{l cccccc}
                          &  \multicolumn{3}{c}{AS-20K (mAP$\uparrow$)} &  \multicolumn{3}{c}{VGGSound (Acc.$\uparrow$)}  \\
                          \cmidrule(l){2-7} 
        Method             & A &V&A+V & A &V&A+V \\
        
        \toprule
          \ourmodel~(w/ AST)                                       & \bf37.5   & 9.6 & 38.1   &\bf 61.3 &30.4 &62.5       \\
          \ourmodel~(w/ ViT)                                          & 36.5   & \bf23.7 & \bf41.6   & 55.7 &\bf46.0&\bf64.9       \\
          %
        \bottomrule
    \end{tabular}
}
\vspace{\tabmargin}
\end{table*}

\begin{table*}[t]
\caption{
    \textbf{Generalization to Audio-only and Visual-only Inputs.} 
    We compare the performance of \ourmodel and CAV-MAE in audio-only and video-only input settings. 
    \ourmodel presents the adaptability in dealing with diverse inputs and absent modalities.
    }
    \vspace{\tabmargin}
\centering
\setlength{\tabcolsep}{6pt} %
\resizebox{0.6\linewidth}{!}{
\begin{tabular}{lcccc}
& \multicolumn{2}{c}{\makecell{AS-2M (mAP$\uparrow$)}} & \multicolumn{2}{c}{\makecell{VGGSound (Acc.$\uparrow$)}} \\
\cmidrule(l){2-5} 
Method      & A & V & A& V \\ \midrule
CAV-MAE                        & 43.3 & 11.1 & 51.8 & 27.3 \\  
\ourmodela         & \bf45.2 & \bf30.8  & \bf55.7 & \bf46.0 \\ 
\bottomrule
\label{tab:robust}
\end{tabular}
}
\vspace{-0.4cm}
\end{table*}

\begin{table}[t]
    \caption{
    \textbf{Importance of Pretraining Objectives.}
    We analyze the impact of various pretraining objectives on downstream audio-visual classification performance. We observe that pretraining with a joint contrastive and reconstruction loss ($\mathcal{L}_{c} + \mathcal{L}_{rec}$) achieves the best results.
    }
    \vspace{\tabmargin}
    \label{tab:cl_and_mae}
    \centering
    \resizebox{0.65\textwidth}{!}{
    \begin{tabular}{l cccc}
        \toprule
        Method    & V100 Hours & \makecell{AS-20K} & \makecell{VGGSound}\\
        \toprule
         No Self-Supervised Pretraining             & - & 39.9    & 61.2 \\

           AV Contrastive Learning ($\mathcal{L}_c$)                              & \bf 160& 41.3    & 64.4 \\
           \ourmodela-Base ($\mathcal{L}_{c} + \mathcal{L}_{rec}$)                                      & 177& \bf41.6  & \bf64.9 \\
        \bottomrule
    \end{tabular}
}
\vspace{\tabmargin}
\end{table}

\subsection{Ablation Studies}
\vspace{\subsecmargin}
\label{sec:ablation}
 Next, we present our ablation studies to (1) validate the effectiveness of our proposed multi-ratio scheme, (2) compare our model with an equivalent separate-encoder baseline, (3) study the effect of using audio vs visual encoder as our shared backbone, (4) assess the generalization of our model to audio-only and video-only settings, and lastly, (5) investigate the importance of unsupervised audio-visual pretraining objectives.

\textbf{The Effectiveness of a Multi-Ratio Scheme.}
In \tabref{ratio}, we compare the fixed ratio masking baselines with our multi-ratio scheme using two metrics: (1) mAP audio-visual classification accuracy on AudioSet-20K and (2) the pretraining cost on AudioSet-2M (in V100 GPU hours). All variants in the Table are first pretrained on AudioSet-2M and then finetuned on AudioSet-20K. 
Based on these results, we first note that the multi-ratio masking scheme variant outperforms all fixed-ratio masking variants.
In particular, our multi-ratio masking outperforms the best fixed-ratio masking scheme (\textbf{41.3} mAP vs. \textbf{40.8} mAP) while being significantly cheaper to pretrain (\textbf{160} vs. \textbf{362} V100 GPU hours).
Furthermore, our multi-ratio masking approach also outperforms the standard contrastive learning baseline (\textbf{41.3} vs \textbf{40.4} mAP) while also being cheaper to pretrain (\textbf{160} vs. \textbf{510} V100 GPU hours).
Next, we observe that although the most efficient fixed-ratio masking scheme needs fewer pretraining hours (\ie \textbf{120} and \textbf{160} V100 hours), it also performs much worse than our multi-ratio masking scheme (\ie \textbf{38.6} mAP and \textbf{41.3} mAP).
Thus, these results suggest that our multi-ratio input masking improves performance while being more efficient than the best-performing fixed-ratio masking or contrastive learning baselines.

\textbf{Comparison with a Separate-encoder Baseline.} In \tabref{abs_gpu}, we also compare our \ourmodela~model with a baseline that uses separate audio and visual encoders (referred to as AVSep). Both baselines use the same implementation, except that our model uses a shared audio-visual encoder, whereas AVSep uses separate audio and visual encoders. Our results indicate that despite a significantly smaller model capacity (\textbf{1.92$\times$} fewer parameters), our \ourmodela~achieves comparable or even slightly better performance than the variant with separate audio and visual backbones on AudioSet-2M. Our model also requires significantly less GPU memory (\textbf{10.9G} vs. \textbf{20.6G}).


\begin{figure*}[t]
    \centering
	\includegraphics[width=0.9\linewidth]{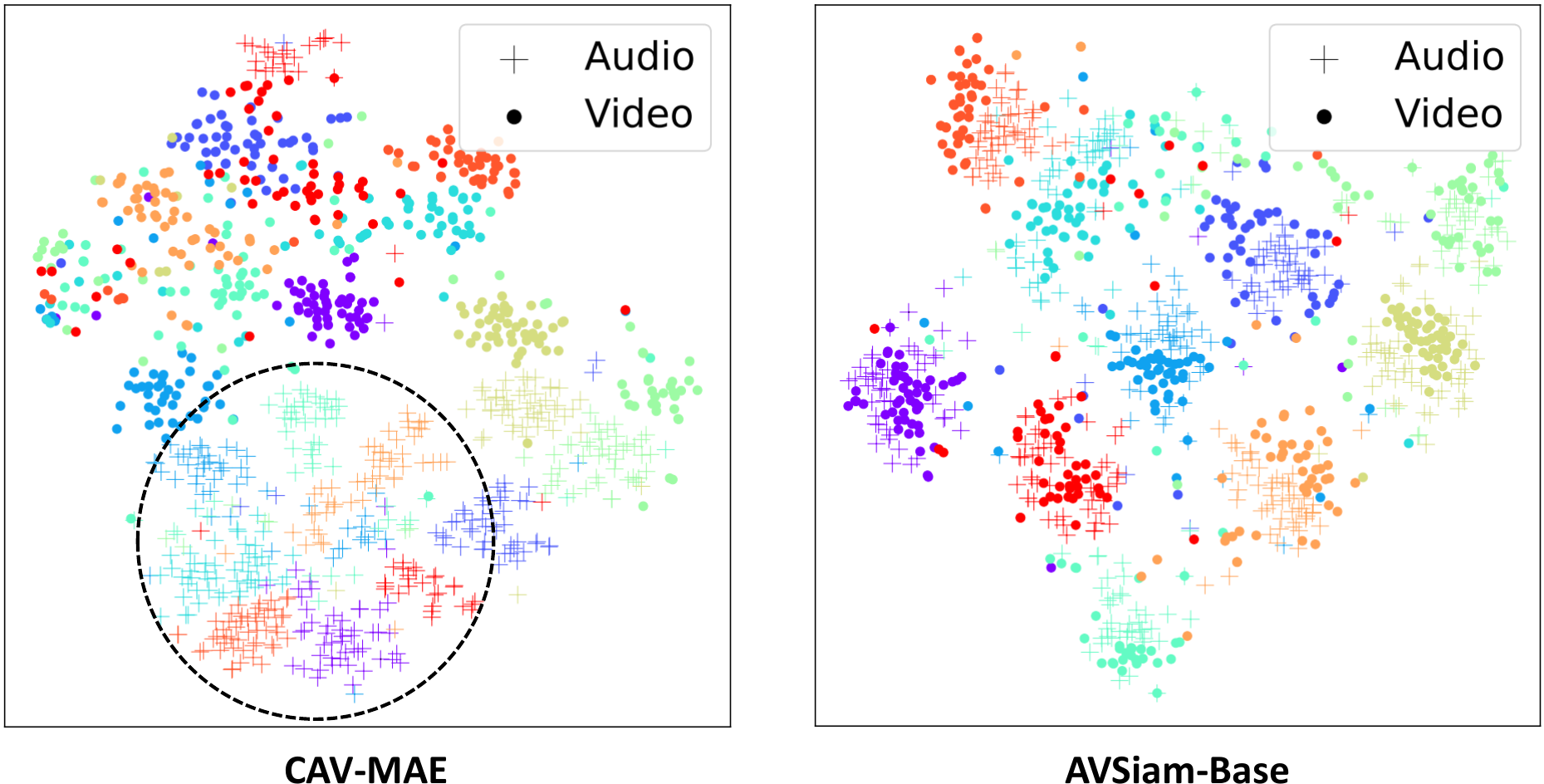}
    \caption{
    \textbf{t-SNE Audio and Image Embedding Visualization.} We use t-SNE to visualize the audio and visual features extracted by (1) a baseline that uses separate audio and visual encoders (\ie CAV-MAE) and (2) our shared-weight encoder method (\ie \ourmodela-Base) on the VGGSound dataset. Each point in the plot represents a single input ($+$ for audio and \protect\tikz \protect\draw[black,fill=black] (0,0) circle (.5ex); for visual), while different colors depict distinct audio-visual categories. Based on this illustration, we observe that \ourmodela~learns more semantically separable features than CAV-MAE. Furthermore, unlike CAV-MAE, \ourmodela~groups audio and visual features corresponding to the same audio-visual category into the same clusters. This suggests that compared to the methods that use separate audio and visual encoders, our \ourmodela~with a shared-weight audio-visual encoder learns to encode audio and visual features into a more similar latent space. 
}
	\label{fig:tsne}
\vspace{\figmargin}
\end{figure*}

\textbf{Using ViT vs. AST as a Shared Encoder.}  In \tabref{rebuttal_ast}, we also compare our default approach that uses ViT as its shared audio-visual encoder with a variant that uses an Audio Spectrogram Transformer (AST)~\cite{interspeech21_AST} instead of a ViT. Based on these results, we observe that the AST-based variant achieves better results for audio modeling (\ie \textbf{+1} mAP on AudioSet-20K and \textbf{+5.6\%} on VGGSound) but significantly worse results for visual and audio-visual classification on AudioSet-20K (\textbf{-14.1} and \textbf{-3.5} mAP) and VGGSound (\textbf{-15.6}\% and \textbf{-2.4}\%). Thus, we use ViT as our shared audio-visual backbone.

\textbf{Generalization to Audio-only and Visual-only Inputs.}
In \tabref{robust}, we evaluate our \ourmodela~model and CAV-MAE on audio-only and visual-only inputs. Our results suggest that our model produces significantly better results than CAV-MAE in these settings.
In particular, in the audio-only setting, our model outperforms CAV-MAE on AudioSet-2M (\ie \textbf{45.2} mAP vs. \textbf{43.3} mAP) and VGGSound (\ie \textbf{55.7\%} vs. \textbf{51.8\%}).
Furthermore, in the visual-only input setting, the performance gap between CAV-MAE and \ourmodela is even larger on both AudioSet-2M (\ie \textbf{30.8} mAP vs. \textbf{11.1} mAP), and VGGSound (\ie \textbf{46.0\%} vs. \textbf{27.3\%}).
These results show the flexibility of our model to handle variable inputs and missing modalities (e.g., visual-only, audio-only, audio-visual inputs).

\textbf{Importance of Pretraining Objectives.} 
Lastly, we examine the significance of our pretraining objectives. 
In \tabref{cl_and_mae}, we report the downstream audio-visual classification accuracy on the AudioSet-20K and VGGSound datasets.
We observe that directly performing supervised fine-tuning (i.e., no unsupervised pretraining) leads to significantly worse performance on both AudioSet-20K (\textbf{-1.7} mAP) and VGGSound (\textbf{-3.7\%}), compared to our default variant that uses unsupervised pretraining.
Pretraining with an audio-visual contrastive objective only ($\mathcal{L}_c$) leads to a boost in performance (\ie \textbf{+1.4} mAP on AudioSet-20K and \textbf{+3.2\%} on VGGSound). 
Furthermore, incorporating the MAE objective in the multimodal decoder ($\mathcal{L}_{rec}$) produces a slight improvement (\textbf{+0.3} mAP and \textbf{+0.5\%}) with a small increase in the computational cost (\textbf{+17} V100 hours). 
%
%
%
\subsection{Qualitative Results}
\vspace{\subsecmargin}

Lastly, in \figref{tsne}, we visualize audio and visual features embeddings extracted using (1) a baseline that uses separate audio and visual encoders (\ie CAV-MAE~\cite{cavmae}) and (2) our shared-weight encoder approach (\ie \ourmodela-Base). The visualization is done using t-SNE~\cite{tsne} on the VGGSound dataset. Each point in the plot represents a single input ($+$ for audio and \tikz\draw[black,fill=black] (0,0) circle (.5ex); for visual), while different colors depict distinct audio-visual categories.
Based on this visualization, we first observe that our \ourmodela~produces semantically more separable features than  CAV-MAE.
Additionally, we observe that, unlike CAV-MAE, \ourmodela~groups audio and visual features corresponding to the same audio-visual category into the same clusters.
This suggests that compared to the approaches that use separate audio and visual encoders, our method encodes audio and visual features into a more similar latent space, which is particularly helpful for tasks such as audio-visual retrieval, as demonstrated by our quantitative results above.

\vspace{\secmargin}
\section{Conclusions}
\vspace{\secmargin}
\label{sec:conclusions}

In this paper, we propose \ourmodela, a framework that uses a single shared ViT encoder for audio and visual data and a novel multi-ratio masking scheme for efficient audio-visual pretraining.
\ourmodela~is fast, scalable, and memory-efficient, and it achieves state-of-the-art results on multiple audio-visual classification and retrieval datasets. In the future, we plan to leverage the efficiency and scalability of our \ourmodela~model and scale it to even larger datasets and model sizes. We will also extend our framework to other audio-visual understanding tasks, such as audio-visual question-answering, event localization, and segmentation.

\section*{Acknowledgments}
 We thank Feng Cheng, Md Mohaiminul Islam, Ce Zhang, Yue Yang, and Soumitri Chattopadhyay for their helpful discussions. This work was supported by the Sony Faculty Innovation Award, Laboratory for Analytic Sciences via NC State University, ONR Award N00014-23-1-2356.

\newcount\cvprrulercount
\appendix
\begin{table*}[t]
\centering
\caption{\textbf{Hyper-parameter for \ourmodela-Base.}}
\label{tab:hyperpar_base}
\resizebox{0.9\textwidth}{!}{
\begin{tabular}{lcccc}
\toprule
                       & \multicolumn{4}{c}{\ourmodela-Base }       \\ 
                       \midrule
                       & \multicolumn{1}{c}{Pretraining} & \multicolumn{3}{c}{Finetuning} \\ 
                       \cmidrule(l){2-5} 
Dataset                & AS-2M                    & AS-20K   & AS-2M   & VGGSound  \\
Optimizer              & \multicolumn{4}{c}{Adam, weight decay=5e-7, betas=(0.95, 0.999)} \\
Backbone learning rate & 1e-4                     & 1e-4     & 5e-6    & 5e-5      \\
LR Classifier ($\times$ encoders)  & -             & 100     & 50    & 10      \\
LR decay start epoch   & 10                        & 2        & 5       & 2         \\
LR decay rate          & 0.5                        & 0.75      & 0.75     & 0.75       \\
LR decay step          & 5                          & 1        & 1       & 1         \\
Epochs                 & 20                         & 15       & 15      & 15        \\
Batch size per GPU     & 96                         & 8       & 32      & 32        \\
GPUs                   & 16 $\times$ A5000   & \multicolumn{3}{c}{4 $\times$ A5000}               \\
Class Balance Sampling & No                          & No       & Yes     & Yes       \\
Mixup                  & No                         & Yes      & Yes     & Yes       \\
Random Time Shifting   & Yes                       & Yes      & Yes     & Yes       \\
Loss Function          & CL + MAE           & BCE      & BCE     & CE        \\
Input Norm Mean        & -5.081                & -5.081   & -5.081  & -5.081    \\
Input Norm STD         & 4.485                   & 4.485    & 4.485   & 4.485     \\ \bottomrule
\end{tabular}
}
\end{table*}

\section*{Appendix}
\section{Implementation Details}
For all of our experiments, the video length is set to $10$ seconds.
During the pretraining stage, we randomly select one video frame with a spatial resolution of $224 \times 224$ from all available video frames.
For audio preprocessing, we first resample the audio waveform to $16$ kHz and then compute the audio spectrogram using PyTorch's~\cite{pytorch} kaldi fbank. This process includes $128$ triangular mel-frequency bins and a frameshift of $10$ milliseconds.
A $10$-second audio spectrogram has a spatial resolution of $1024 \times 128$.
We follow the standard ViT pipeline, which first patchifies the data and then computes the self-attention mechanism for both audio and visual tokens.
For the MAE loss, we adopt the same loss function as used in CAV-MAE~\cite{cavmae}, applying the stopping gradient before the multimodal layers.
During fine-tuning, similar to CAV-MAE~\cite{cavmae}, we aggregate all predicted probabilities across all video frames (\ie 10 video frames) to make more accurate predictions on video data. 
This approach may be less effective than the video encoders used in AV-MAE~\cite{avmae} and MAViL~\cite{nips23_mavil}.
We include all the detailed hyper-parameter settings in \tabref{hyperpar_base}, \tabref{hyperpar_large}, and \tabref{hyperpar_huge}.

\begin{table*}[h]
\centering
\caption{\textbf{Hyper-parameter for \ourmodela-Large.}}
\label{tab:hyperpar_large}
\resizebox{0.9\textwidth}{!}{
\begin{tabular}{lcccc}
\toprule
                       & \multicolumn{4}{c}{\ourmodela-Large }       \\ 
                       \midrule
                       & \multicolumn{1}{c}{Pretraining} & \multicolumn{3}{c}{Finetuning} \\ 
                       \cmidrule(l){2-5} 
Dataset                & AS-2M                    & AS-20K   & AS-2M   & VGGSound  \\
Optimizer              & \multicolumn{4}{c}{Adam, weight decay=5e-7, betas=(0.95, 0.999)} \\
Backbone learning rate & 1e-4                     & 5e-5     & 5e-6    & 5e-6      \\
LR Classifier ($\times$ encoders)  & -             & 100     & 50    & 50      \\
LR decay start epoch   & 10                        & 2        & 2      & 2         \\
LR decay rate          & 0.5                        & 0.75      & 0.5     & 0.5       \\
LR decay step          & 5                          & 1        & 1       & 1         \\
Epochs                 & 20                         & 15       & 15      & 15        \\
Batch size per GPU     & 48                         & 8       & 16      & 16        \\
GPUs                   & 32 $\times$ A5000   & \multicolumn{1}{c}{4 $\times$ A5000}  & \multicolumn{2}{c}{8 $\times$ A5000}               \\
Class Balance Sampling & No                          & No       & Yes     & Yes       \\
Mixup                  & No                         & Yes      & Yes     & Yes       \\
Random Time Shifting   & Yes                       & Yes      & Yes     & Yes       \\
Loss Function          & CL + MAE           & BCE      & BCE     & CE        \\
Input Norm Mean        & -5.081                & -5.081   & -5.081  & -5.081    \\
Input Norm STD         & 4.485                   & 4.485    & 4.485   & 4.485     \\ \bottomrule
\end{tabular}
}
\end{table*}

\begin{table*}[b]
\centering
\caption{\textbf{Hyper-parameter for \ourmodela-Huge.}}
\label{tab:hyperpar_huge}
\resizebox{0.9\textwidth}{!}{
\begin{tabular}{lcccc}
\toprule
                       & \multicolumn{4}{c}{\ourmodela-Huge }       \\ 
                       \midrule
                       & \multicolumn{1}{c}{Pretraining} & \multicolumn{3}{c}{Finetuning} \\ 
                       \cmidrule(l){2-5} 
Dataset                & AS-2M                    & AS-20K   & AS-2M   & VGGSound  \\
Optimizer              & \multicolumn{4}{c}{Adam, weight decay=5e-7, betas=(0.95, 0.999)} \\
Backbone learning rate & 1e-4                     & 1e-5     & 5e-6    & 5e-6      \\
LR Classifier ($\times$ encoders)  & -             & 100     & 50    & 50      \\
LR decay start epoch   & 10                        & 5        & 2      & 2         \\
LR decay rate          & 0.5                        & 0.5      & 0.5     & 0.5       \\
LR decay step          & 5                          & 1        & 1       & 1         \\
Epochs                 & 20                         & 15       & 15      & 15        \\
Batch size per GPU     & 24                         & \multicolumn{3}{c}{4}        \\
GPUs                   & 64 $\times$ V100           & \multicolumn{3}{c}{8 $\times$ A5000}               \\
Class Balance Sampling & No                          & No       & Yes     & Yes       \\
Mixup                  & No                         & Yes      & Yes     & Yes       \\
Random Time Shifting   & Yes                       & Yes      & Yes     & Yes       \\
Loss Function          & CL + MAE           & BCE      & BCE     & CE        \\
Input Norm Mean        & -5.081                & -5.081   & -5.081  & -5.081    \\
Input Norm STD         & 4.485                   & 4.485    & 4.485   & 4.485     \\ \bottomrule
\end{tabular}
}
\end{table*}

\clearpage

%
%
\bibliographystyle{ieee_fullname}
\bibliography{main}
\end{document}